%% file: 0-main.tex
\documentclass[letterpaper, 10 pt, journal]{ieeeconf}

\IEEEoverridecommandlockouts                    

\usepackage{chemformula}
\title{\LARGE \bf
Control and Morphology Optimization of \\Passive Asymmetric Structures for Robotic Swimming}

\author{Nana Obayashi*$^{1}$, Andrea Vicari*$^{1,2}$, Kai Junge$^{1}$, Kamran Shakir$^{1,3}$, and Josie Hughes$^{1}$
\thanks{*These authors contributed equally to this work}
\thanks{$^{1}$CREATE Lab, EPFL, Lausanne, Switzerland. $^{2}$Scuola Superiore Sant’Anna, Pisa, Italy. $^{3}$Wroc\l{}aw University of Science and Technology, Wroc\l{}aw, Poland. Contact emails: {\tt\small nana.obayashi@epfl.ch, andrea.vicari@santannapisa.it, kai.junge@epfl.ch, josie.hughes@epfl.ch}.}%
}

\begin{document}

\maketitle
\thispagestyle{empty}
\pagestyle{empty}

\begin{abstract}

Aquatic creatures exhibit remarkable adaptations of their body to efficiently interact with the surrounding fluid. The tight coupling between their morphology, motion, and the environment are highly complex but serves as a valuable example when creating biomimetic structures in soft robotic swimmers.
We focus on the use of asymmetry in structures to aid thrust generation and maneuverability. Designs of structures with asymmetric profiles are explored so that we can use morphology to `shape' the thrust generation. 
We propose combining simple simulation with automatic data-driven methods to explore their interactions with the fluid. 
The asymmetric structure with its co-optimized morphology and controller is able to produce 2.5 times the useful thrust compared to a baseline symmetric structure. Furthermore these asymmetric feather-like arms are validated on a robotic system capable of forward swimming motion while the same robot fitted with a plain feather is not able to move forward.


\end{abstract}

\section{Introduction}
\input{1-intro}

\section{Methods}
\input{2-methods}

\section{Results}
\input{3-results}

\section{Discussion \& Conclusion}
\input{4-discussion}





\section*{Acknowledgment}
This project was partially supported by the EU's Horizon 2020 research and innovation program under the Marie Skłodowska Curie grant agreement N$^\circ$ 945363.

\bibliographystyle{IEEEtran}
\bibliography{references}

\end{document}

%% file: 1-intro.tex
\label{sec:intro}
Asymmetry or directionality is a property widely exploited in robotics; from asymmetric friction profiles to enable locomotion~\cite{xu2022locomotion,zhu2017architectures}, asymmetric weight distribution for passive walking~\cite{zou2006effect}, asymmetric structures to enable turning~\cite{shepherd2011multigait}, or asymmetric control of robots~\cite{jia2015energy,thandiackal2021emergence}. Biology shows further examples of where passive or active asymmetry in structure is exploited for advantageous  properties. Aquatic creatures are particularly adept at adjusting or utilizing their body structure or properties to aid their interactions with fluids, enabling complex behaviors to emerge from simple motion patterns.  One such animal that exploits asymmetry in structure to aid thrust generation and maneuverability is a marine crinoid called the `feather star'~\cite{feather_video}.  These animals can alter their limb geometry for asymmetrical thrust generation depending on its desired movement.  Inspired by this use of geometry to `shape' the thrust generation, soft robotic swimmers that use passive structures that break or change their symmetry could similarly utilize structure change to aid their motion.  

Previous work has shown that asymmetric actuators that utilize folding and bending can be used to obtain net positive displacement~\cite{wu2019symmetry-breaking}. Another work has qualitatively investigated an efficient stroke pattern to produce unidirectional thrust with similar passive structures~\cite{chen2018controllable}. However, the optimization and between the structure and the control input was not explored.
In order to exploit passive asymmetry in thrust generating structures that results from passive structures in fluids, we must understand the tight coupling between the dynamic motion, or control and the passive morphology of the asymmetric profile, and  the resultant thrust. This requires high fidelity and accurate modeling of both the fluid-soft structure~\cite{huang2021modeling} and also the large deformations caused by passive properties of the asymmetric structure.  Fluid-solid interaction methods such as the immersed-boundary method are popular for modeling biological systems with large active or passive deformations~\cite{wang2019numerical}, but are still subject to large reality gaps and requires high computational power~\cite{tian2014fluid}. Data-driven model-based approaches are showing potential by combining simulation with experimental methods of passive soft structures in water as a way of reducing the reality gap~\cite{obayashi2022soft}. To develop robots that are optimized to exploit this asymmetry, we must develop methods to explore accurately the relationship between the design of the asymmetric structure and the input controller.  

\begin{figure}[tb]
    \centering
    \includegraphics[width=\linewidth]{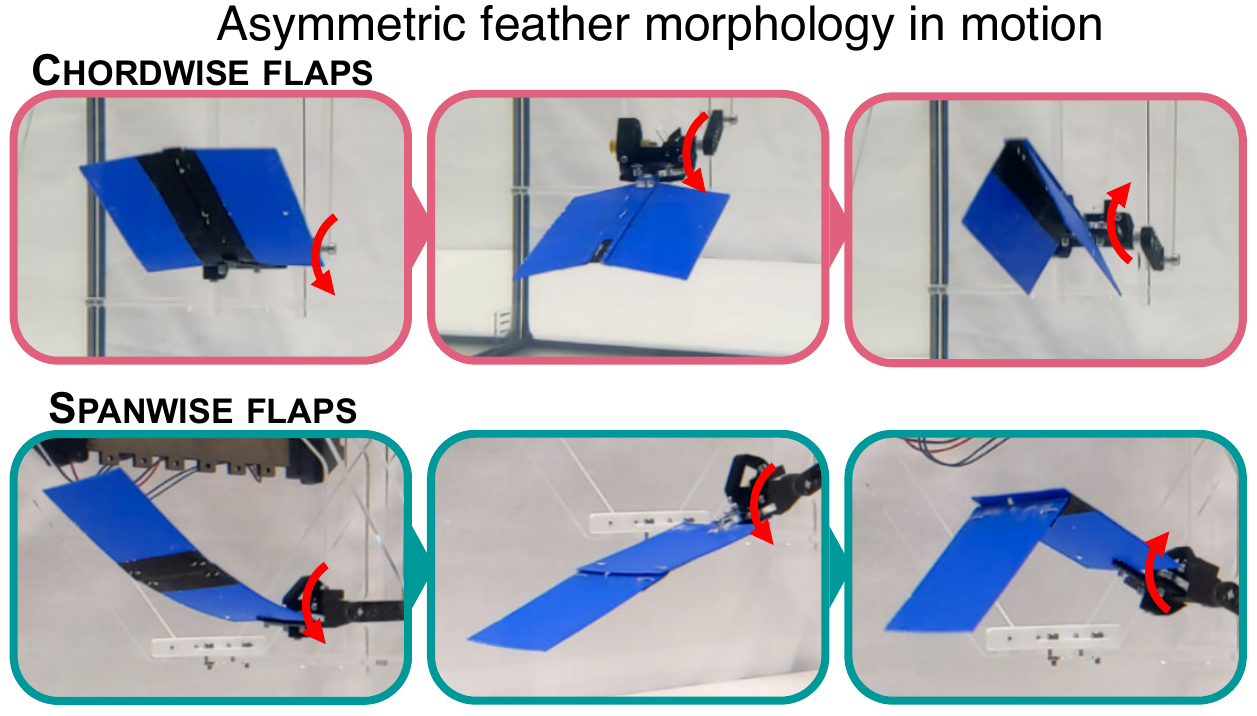}
    \caption{Two designs of asymmetric feather morphology while flapping in a periodic motion. Time lapses of feathers with flaps flapping in the chordwise (\textbf{Left}) and spanwise (\textbf{Right}) directions.}
    \label{fig:fig1}
    \vspace{-4mm}
\end{figure}

\begin{figure}[tb]
    \centering
    \includegraphics[width=1\linewidth]{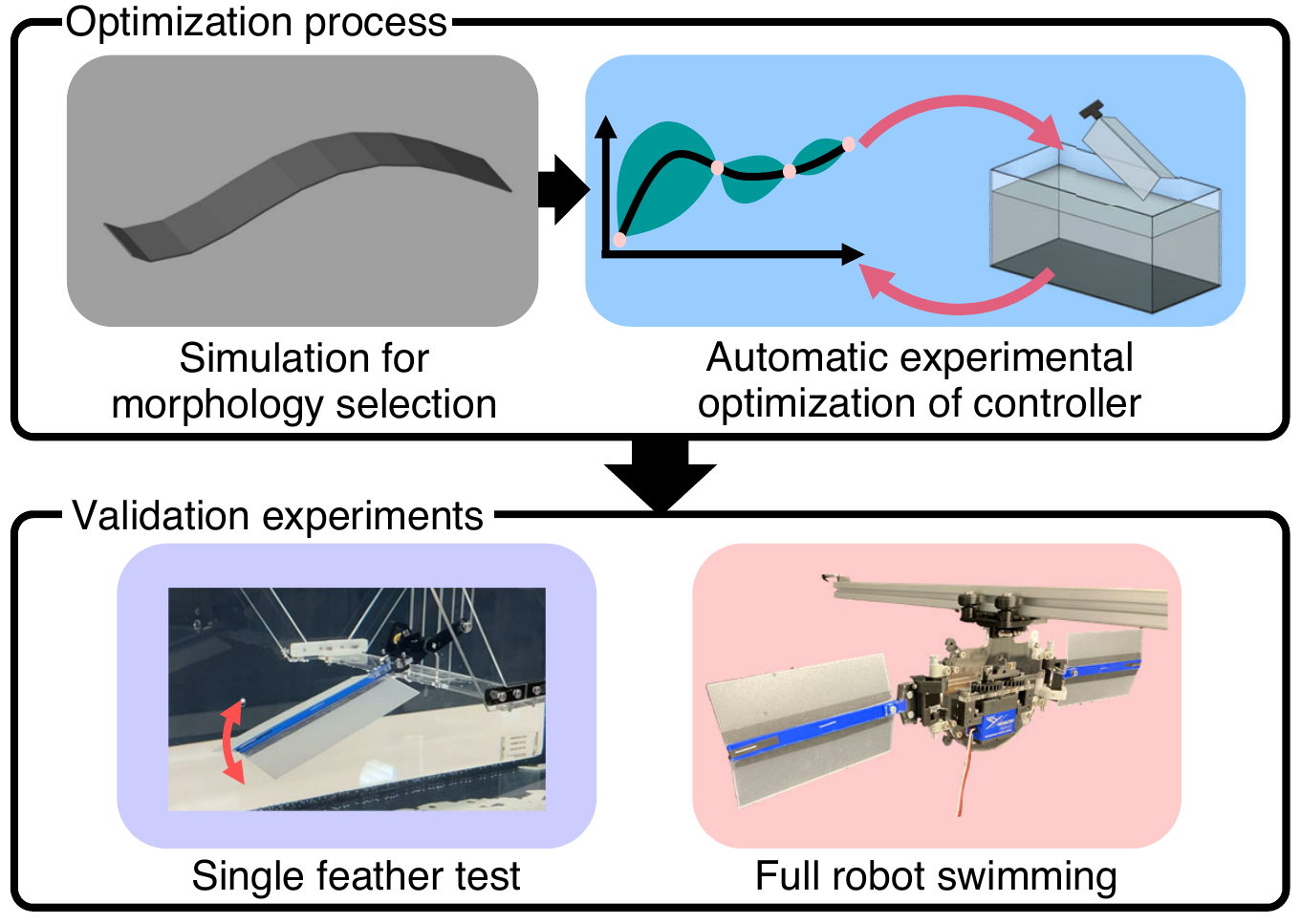}
    \caption{Summary of our approach to co-optimize the morphology and controller by using simulation for preliminary morphology selection followed by an iterative automatic experimental optimization to find the best controller. Validation experiments are then performed.}
    \label{fig:approach}
    \vspace{-4mm}
\end{figure}

By finding an optimal controller and morphology of the feather that exploits asymmetry, we predict that the structure can be used to generate useful directionality in thrust compared to a plain feather without any asymmetric profile. 
However, since the soft body-fluid interactions are challenging to predict with methods that are computationally expensive, we propose utilizing a combination of simulation and data-driven approaches to find the optimal solution.
Driven by bio-inspiration, we explore different designs of asymmetric feathers (Fig.~\ref{fig:fig1}) that utilize foldable `flaps' to create a passive structure which show asymmetric thrust profiles when actuated in the correct conditions in water.  
In the downstroke, the feather flaps lay flat against the resistance of the water due to the limited rotation by the feather spine. In the upstroke, the feather flaps are compliant to the flow as there is no rotational limitation provided by the spine. This passive mechanism allows for an asymmetric profile in the down- and upstrokes similar to the feather contractions of the biological feather star. 
As shown in Fig.~\ref{fig:approach}, using simulation we identify target morphologies of these feather structures which should then be investigated experimentally to find an optimal control input to best exploit the morphology.  By developing an autonomous test bed that performs online optimization of the controller by measuring the thrust, determining an objective function and searching the control space guided by Bayesian optimization, we identify the optimal controller for a range of different morphologies of feathers.  We finally validate the identified structures on a feather star inspired robot system, to explore how the results from the test-bed translate to a full robotic system.

By automatically exploring the design space of a feather actuation, we demonstrate that asymmetric thrust can be leveraged on robotic hardware.
In the remainder of the paper, we first present methods to address the problem. Feather design is shown, followed by experimental results. We will discuss the results after which we will conclude with suggestions for future work.

%% file: 2-methods.tex
To achieve motion through asymmetric thrust, we wish to maximize the difference in thrust between the up- and downstrokes produced by the feather. 
The thrust is generated through the feather-fluid interactions, which are dependent on the morphology of the feather actuated in a periodic motion at the root. 
The feather design parameters are first defined, followed by preliminary investigation of feather morphology using a low-cost hydrodynamics simulation. Taking a subset of feather geometries identified through simulation, the best controller for each is found through a custom physical experimental setup and online iterative optimization.  

\subsection{Feather design for asymmetric thrust}

\begin{figure}[tb]
    \centering
    \includegraphics[width=1\linewidth]{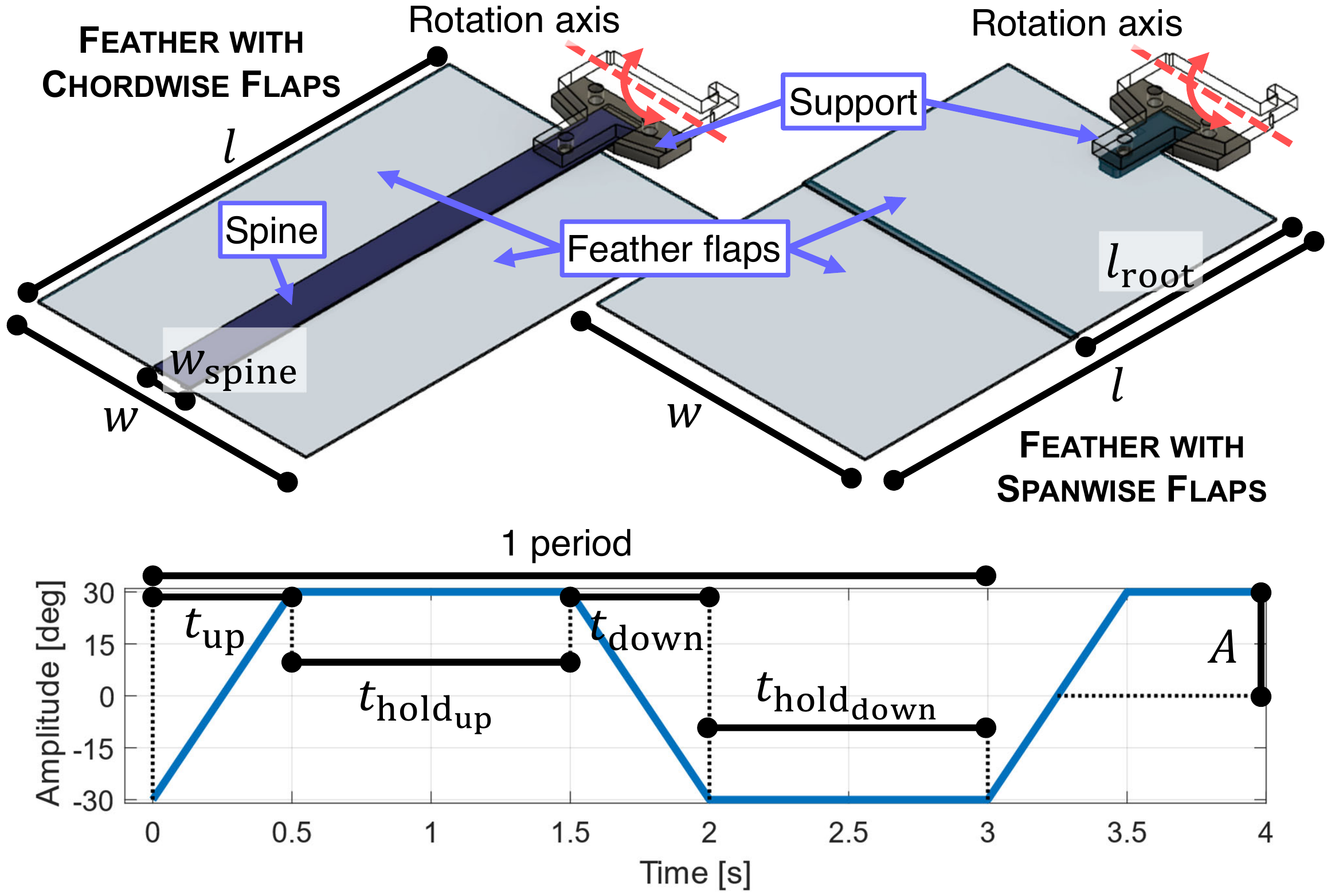}
    \caption{\textbf{Top: }Parameterized feather designs with chordwise flaps and spanwise flaps. \textbf{Bottom: }Parameterized feather motion.}
    \label{fig:asymmetric}
    \vspace{-4mm}
\end{figure}

Feathers with two types of asymmetric flaps are designed, as shown in Figs.~\ref{fig:asymmetric} and \ref{fig:fig1}. When the flaps on the feather are unfolded, the feather has a rectangular shape where it is defined by length, $l$ and width, $w$. The feathers with chordwise flaps have joints on either side of the spine with width, $w_\mathrm{spine}$ which is constant at $1\, \mathrm{cm}$. The feathers with spanwise flaps have a joint at a distance, $l_\mathrm{root}$ from the root where the support is. 
$0.4 \, \mathrm{mm}$-thick polypropylene sheets are used as material for the feathers due to their flexibility and ease of fabrication using a \ch{CO2} laser cutter. Tape is used to attach the feather pieces to enable asymmetric flapping. 

A parameterized controller is explored, which can be described by the rise time, $t_\mathrm{up}$, fall time, $t_\mathrm{down}$, hold time after rise, $t_\mathrm{hold_{up}}$, hold time after fall, $t_\mathrm{hold_{down}}$, and a fixed amplitude, $A$ as shown in Fig.~\ref{fig:asymmetric}. This controller is chosen so the coupling between the asymmetric design and the motion can be exploited, specifically the speed of the strokes and the recovery time for the flaps.

\subsection{Hydrodynamic simulation for design space reduction}

As initial exploration of the design space of the feather, hydrodynamics simulation developed in \cite{stella2022controlling} was used to characterize general trends. A single feather is modeled as a collection of discrete flexible units using Simscape Multibody\textsuperscript{TM}. The lumped-parameter method~\cite{miller2017modeling} allows each unit modeled as a spring-mass-damper system to experience deformation similar to that of a soft structure.

For each flexible unit, the total lumped external force, ${F}_\mathrm{ext}$, consists of gravitational force, ${F}_\mathrm{g}$, buoyancy force, ${F}_\mathrm{b}$, hydrodynamic force, ${F}_\mathrm{hyd}$, and added mass force, ${F}_\mathrm{a}$:

\begin{equation}\label{eqn:forces}
    {F}_\mathrm{ext}={F}_\mathrm{g}+{F}_\mathrm{b}+{F}_\mathrm{hyd}+{F}_\mathrm{a}
\end{equation}

\noindent The actuation at the base of the feather drives the kinematics and deformation dynamics. The simulation captures the nonlinear features of the fluid-structure interactions as a dynamic feedback resulting from the motion of the individual elements. Each simulation was run for a specified number of cycles of the periodic motion, and the average thrust, $\overline{T}$ over one period at the base of the feather is recorded: 
\begin{equation}\label{eqn:thrust}
    \overline{T} = \int_0^{t_\mathrm{period}}T(t)dt \approx \frac{1}{n}\Sigma_nT_t
\end{equation}

The simulation is used to identify a reduced set of morphologies to experimentally investigate the asymmetric thrust. 
For identifying a width ratio, $WR=w/w_\mathrm{spine}$ for the feather with chordwise flaps, we fix the length, $l=120\, \mathrm{mm}$ and explore widths, $w=15$-$120\, \mathrm{mm}$ in $5\, \mathrm{mm}$ increments.
Similarly, to identify a length ratio, $LR=l/l_\mathrm{root}$ for the feather with spanwise flaps, the width is fixed at $w=120\, \mathrm{mm}$ and explore lengths, $l=60$-$180\, \mathrm{mm}$ in $10\, \mathrm{mm}$ increments. The geometries are chosen to be small enough compared to the water tank in the experimental setup to minimize edge effects. With those geometries, we explore all combinations of the motion parameters: $t_\mathrm{up}=0.1,0.5$, $t_\mathrm{down}=0.5,1$, $t_\mathrm{hold}=0,0.5,1$ seconds. 

\begin{figure}[tb]
    \centering
    \vspace{1mm}
    \includegraphics[width=0.9\linewidth]{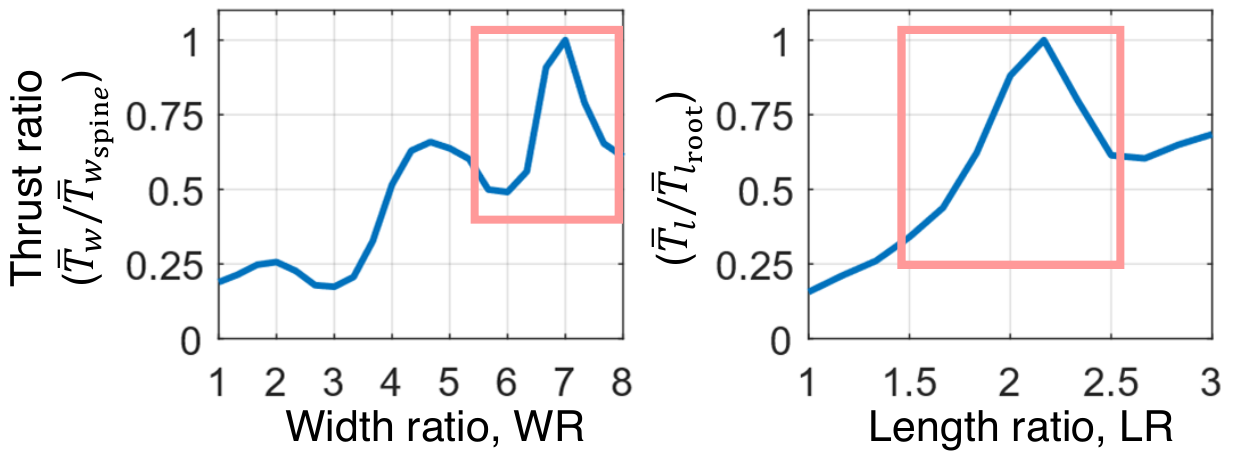}
    \caption{Simulation results showing the thrust ratio for various width and length ratios of the feathers. Thrust ratio is defined as $\overline{T}_\mathrm{w}/\overline{T}_\mathrm{w_{spine}}$ for the feather with chordwise flaps and as $\overline{T}_\mathrm{l}/\overline{T}_\mathrm{l_{root}}$ for the feather with spanwise flaps. A subset of $WR$ and $LR$ is identified with the pink rectangle.} 
    \label{fig:simulation}
    \vspace{-4mm}
\end{figure}

The simulation results for the maximum normalized thrust ratio across all controllers for width ratio and length ratio are shown in Fig.~\ref{fig:simulation}. 
To explore morphologies that provides the highest thrust ratio and therefore highest asymmetric thrust, we explore further experimentally width ratios, $WR=5.5,\ 7.5,\ 9.5$ and length ratios, $LR=1.5,\ 2,\ 2.5$, also identified by the rectangle in Fig.~\ref{fig:simulation}.

\subsection{Experimental setup}

\begin{figure}[tb]
    \centering
    \vspace{1mm}
    \includegraphics[width=1\linewidth]{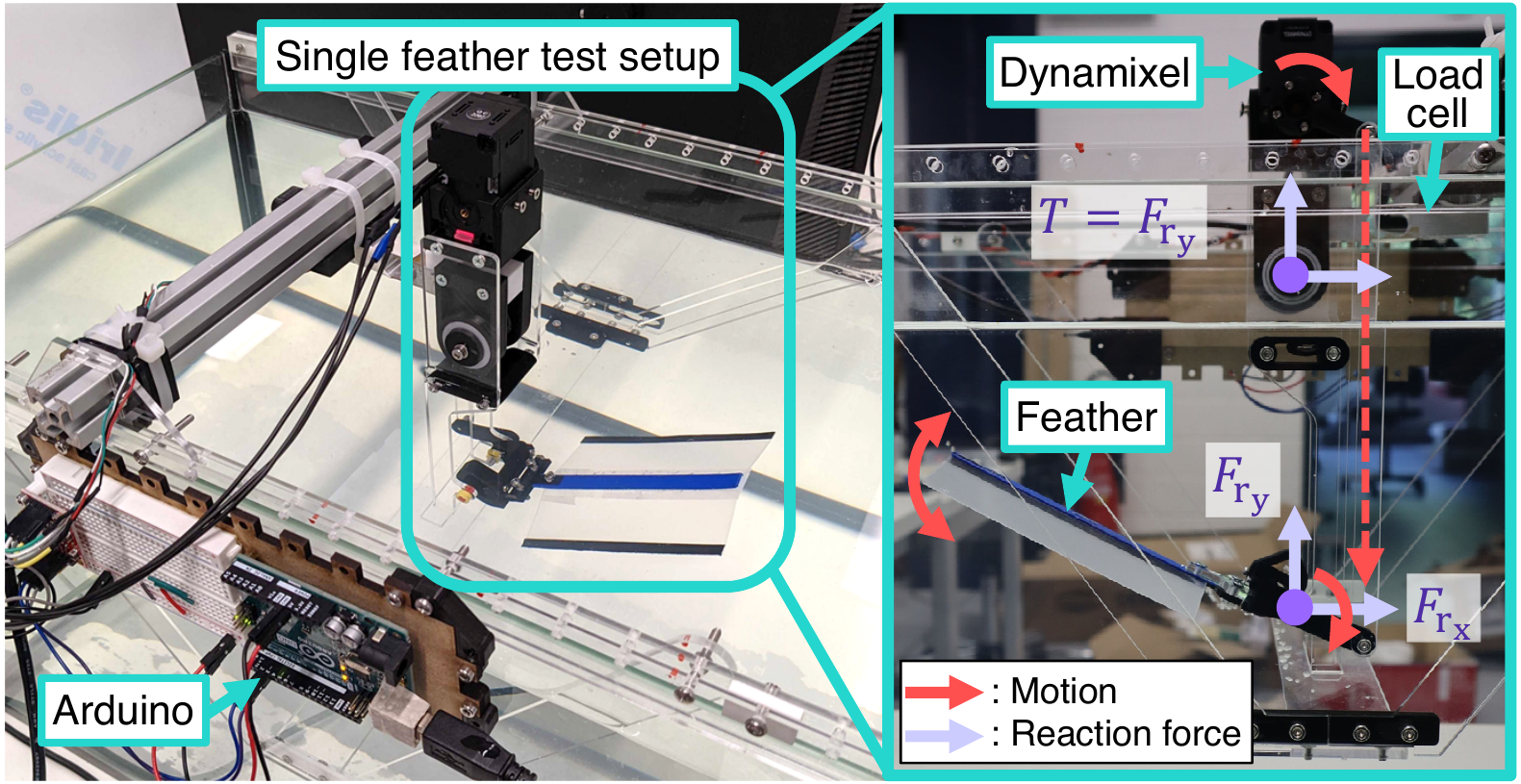}
    \caption{Experimental setup for measuring the thrust generated by feathers.}
    \label{fig:setup}
    \vspace{-4mm}
\end{figure}

To gather experimental data of the generated thrust, an experimental setup (Fig.~\ref{fig:setup}) is created. It uses a servo-powered mechanism to actuate the base of the feather, with a $0$-$3 \, \mathrm{kg}$ load cell used to measure the upwards thrust, $T$. The setup ensures there are no moments applied at the load cell such that it truly measures the upwards thrust. The tank size has been chosen to be significantly larger than the feather to minimize edge effects.
A representative time series obtained from the load cell is shown in Fig.~\ref{fig:res_asymmetric}.

\subsection{Online optimization of feather controller}

For a given geometry of the feather, automated optimization of the controller can be performed using the custom experimental setup. 
As input, we have three control parameter ratios: $\frac{t_\mathrm{up}}{t_\mathrm{down}}$, $\frac{t_\mathrm{hold_{up}}}{t_\mathrm{hold_{down}}}$, and $\frac{t_\mathrm{hold}}{t_\mathrm{move}}$, where $t_\mathrm{hold}=t_\mathrm{hold_{up}}+t_\mathrm{hold_{down}}$ and $t_\mathrm{move}=t_\mathrm{up}+t_\mathrm{down}$. Evaluating control parameters as ratios rather than absolute times allowed for meaningful comparison between the various segments of the feather movement. Furthermore, the logarithms of these ratios are used as optimizable variables for a better sampling distribution. The limits for the control parameters are chosen based on mechanical limits of the servo motor. For each iteration of an experiment, the period of the signal is fixed. 
The single feather in the tank is actuated according to the chosen control parameters for a specified amount of periodic cycles and the thrust data is obtained from the load cell. 

Indeterministic Bayesian optimization~\cite{frazier2018botutorial} is a suitable algorithm to sequentially explore the control parameter design space and to better quantify the uncertainty in the system~\cite{junge2020improving}. 
An exploration ratio of $0.6$ is chosen from heuristic trial and error for the given design problem. 
The objective of the optimization is to find the control parameter ratios that maximize the integral of the upwards to downwards thrust, or thrust ratio, $TR$:

\begin{equation}\label{eqn:thrustratio}
    TR=\frac{\int_0^tT^+(t)dt}{\int_0^t\lvert T^-(t)\rvert dt}
\end{equation}

\noindent where $T^+(t)$ are the thrust datapoints larger than zero and $T^-(t)$ are the thrust datapoints smaller than zero. The thrust ratio quantifies the asymmetricity in thrust during the feather's up- and downstrokes, hence the swimming speed in the upwards direction in the setup (Fig.~\ref{fig:setup}).

\begin{figure}[tb]
    \centering
    \includegraphics[width=0.7\linewidth]{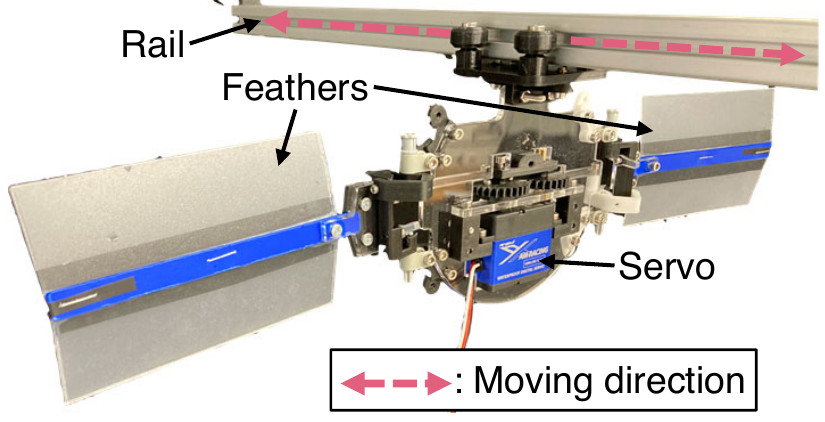}
    \caption{Full two-feather robot platform with labeled components used to perform swimming tests.}
    \label{fig:fullrobot}
    \vspace{-4mm}
\end{figure}

\subsection{Robotic hardware}

To explore how the optimized feather design can be used on a robotic system, two feathers can be connected to a waterproof servo motor using a four-bar linkage mechanism and actuated simultaneously in the same control sequence to create a robotic swimmer. The robot swims along an extruded aluminium rail (Fig.~\ref{fig:fullrobot}) by flapping its two feathers. The robot has a tether to provide power supply and control signal.

%% file: 3-results.tex
\begin{figure}[tb]
    \centering
    \includegraphics[width=1\linewidth]{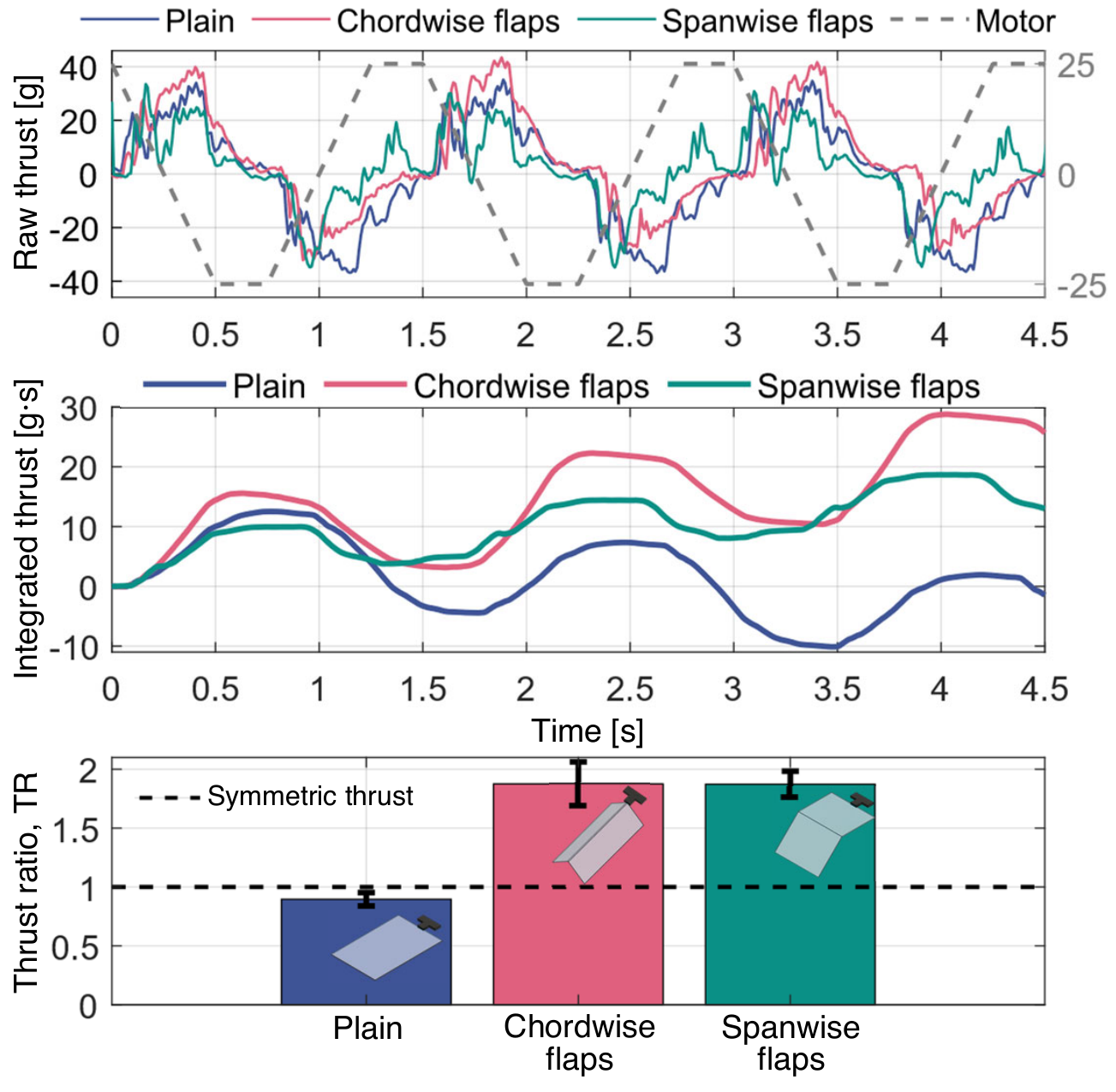}
    \caption{\textbf{Top: }Sample of raw thrust data for a plain feather and feathers with chordwise flaps ($WR=7.5$) and spanwise flaps ($LR=2.0$) flapping with a symmetric motion with a period of $1.5$ seconds. \textbf{Middle: }Integrated raw thrust for the feathers mentioned above, where the feathers with flaps produce a generally upward trending curve providing directional thrust. \textbf{Bottom: }Thrust ratio, $TR$ for the same feathers mentioned above, where feathers with flaps produce almost double the positive thrust compared to negative thrust.}
    \label{fig:res_asymmetric}
    \vspace{-4mm}
\end{figure}

\subsection{Validation of asymmetric feather}
\label{subsec:asymmetric}

Prior to optimizing the controller, we explore different feather morphologies to validate the impact of the asymetric designs.  The different feathers are evaluated using a symmetric controller where $t_\mathrm{down},\ t_\mathrm{up}=0.25\, \mathrm{s}$ and $t_\mathrm{hold_{up}},\ t_\mathrm{hold_{down}}=0.5\, \mathrm{s}$. Three different types of feather are investigated: a plain feather with no flaps which provides a baseline for comparison, a feather with chordwise flaps, and a feather with spanwise flaps. Between the three feathers, the unfolded area is identical: $w=7.5\, \mathrm{cm}$ and $l=11.5\, \mathrm{cm}$. 
The raw thrust data for each feather is shown in Fig.~\ref{fig:res_asymmetric} along with the feather position. In general, during the downstroke, all feathers produce positive thrust and during the upstroke, they produce negative thrust. The integrated thrust is also shown for each feather for better visualizing the asymmetric thrust. 10 sets of thrust data are collected for each feather as they go through five cycles of periodic flapping. Using a symmetric controller, the plain feather produces approximately symmetric thrust ($TR\approx1$) while for the feathers with chordwise and spanwise flaps, the positive thrust is almost double that of negative thrust ($TR\approx2$). The asymmetric thrust we obtain solely from the morphology demonstrates a potential for even larger increase in thrust ratio with an improved controller.

\begin{figure}[tb]
    \centering
    \includegraphics[width=1\linewidth]{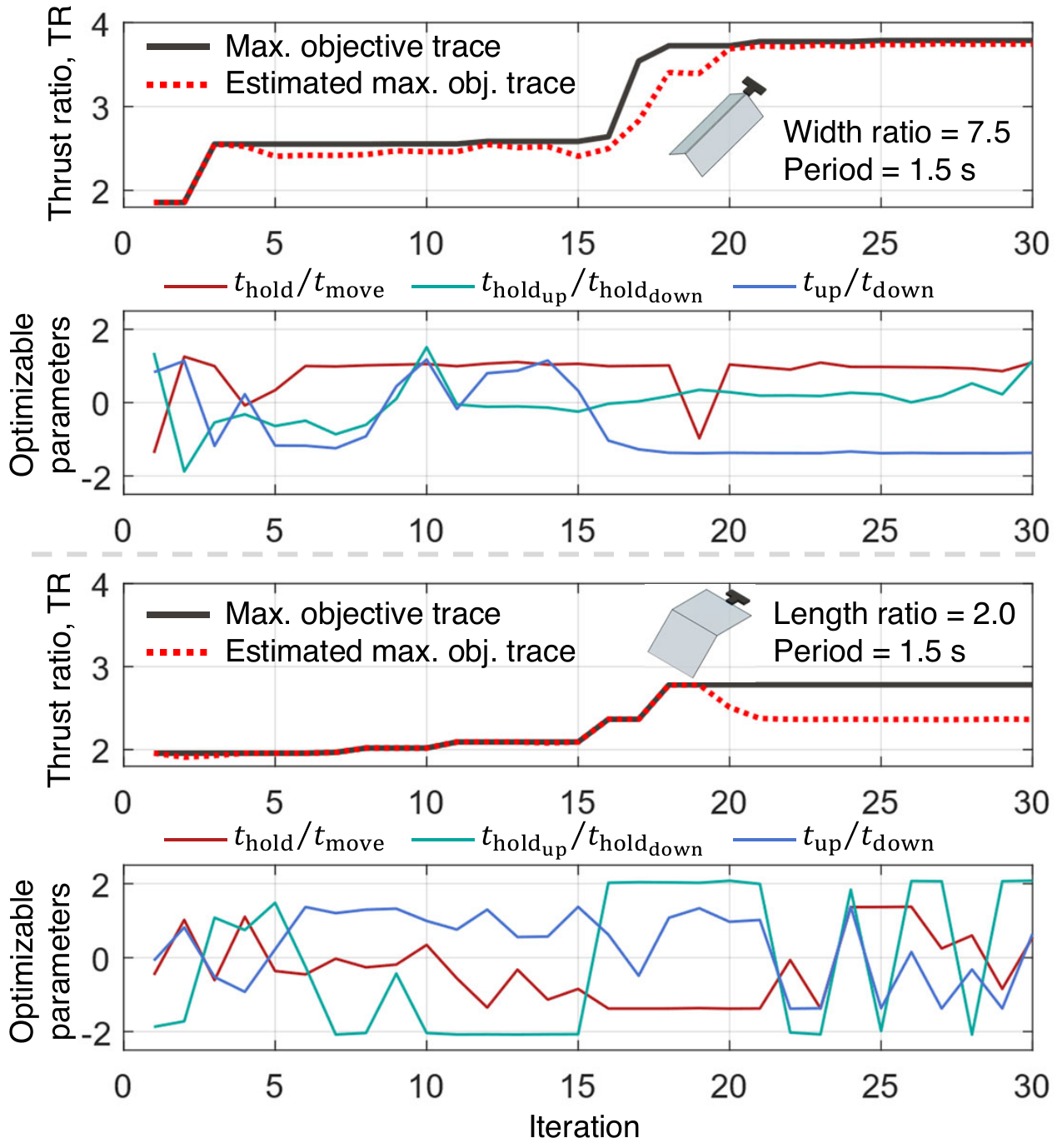}
    \caption{Sample progressions of the Bayesian optimization for the feather motion. The objective is to maximize thrust ratio, $TR$ and the optimizable parameters are the logarithms of time ratios: $t_\mathrm{up}/t_\mathrm{down}$, $t_\mathrm{hold_{up}}/t_\mathrm{hold_{down}}$, and $t_\mathrm{hold}/t_\mathrm{move}$.}
    \label{fig:res_boexample}
    \vspace{-4mm}
\end{figure}

\begin{figure}[tb]
    \centering
    \vspace{1mm}
    \includegraphics[width=1\linewidth]{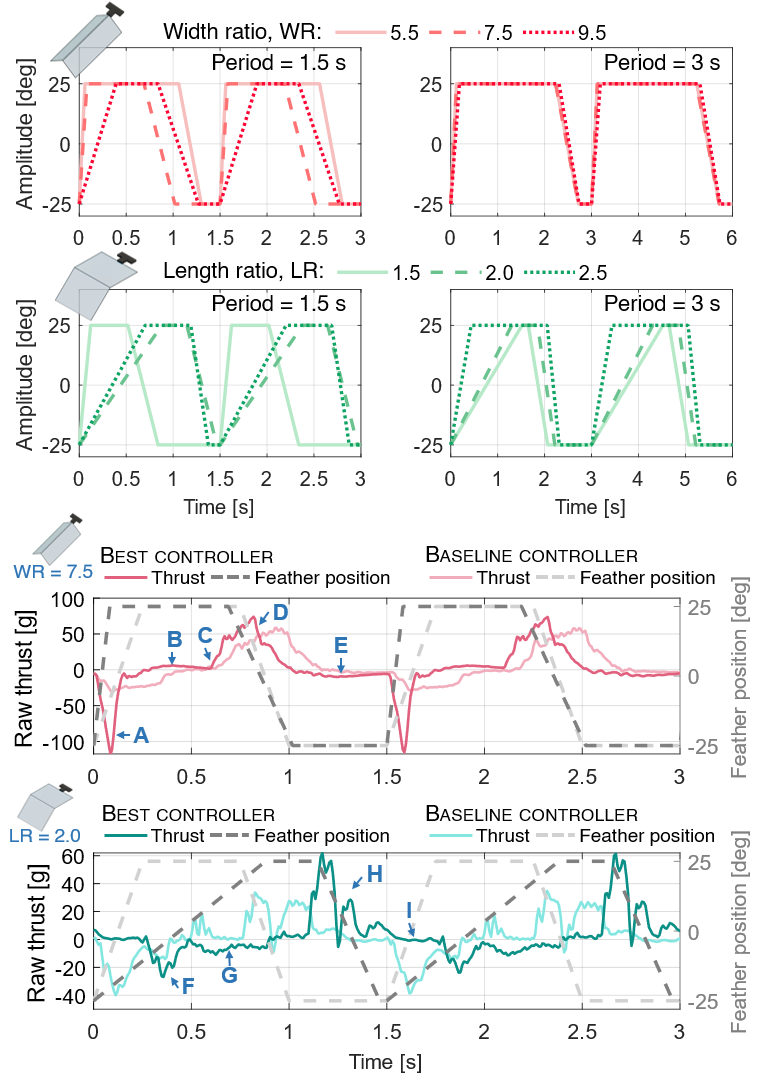}
    \caption{\textbf{Top 2 rows: }Optimized controller for various ratios of feathers with flaps where the period is fixed at $1.5\, \mathrm{s}$ and $3\, \mathrm{s}$. \textbf{Bottom 2 rows: }Example raw thrust data using the best and baseline controllers overlaid with the feather motion with a period of $1.5\, \mathrm{s}$. Thrust data for feather with chordwise flaps with the best controller is characterized by a sharp negative thrust during upstroke (\textbf{A}), recovery periods (\textbf{B} \& \textbf{E}), thrust increase with deployment of flaps (\textbf{C}), and large thrust increase with a full area downstroke (\textbf{D}). Thrust data for spanwise flaps with the best controller is characterized by a delayed negative thrust halfway through the upstroke (\textbf{F}), recovery periods (\textbf{G} \& \textbf{I}), and a series of positive but decreasing peaks (\textbf{H}).}
    \label{fig:res_optcontroller}
    \vspace{-5mm}
\end{figure}

\subsection{Controller optimization for a single feather}

Bayesian optimization is used to find the controller that produces the highest thrust ratio for the three types of feathers--plain, feather with chordwise flaps, and feather with spanwise flaps. Examples of optimization progression and parameters searched are shown as Fig.~\ref{fig:res_boexample} for the feather with chordwise flaps with $WR=7.5$ and one with spanwise flaps with $LR=2.0$ and controller period of $1.5\, \mathrm{s}$. Each experiment is allowed a maximum of 30 iterations, by when the estimated objective value would have converged for this system. 
By iteration 20, the estimated thrust ratio reaches a stable value of $TR\approx4$ for the controller optimization of the feather with chordwise flaps and all parameters converge. 
In the optimization of the feather with spanwise flaps, the optimizer is still exploring the parameters at iteration 30. Nevertheless, with this system having stochastic characteristics, the optimizer still performs well giving a stable estimated thrust ratio $TR\approx2.5$ from iteration 21.

The optimized controller for all tested width and length ratios for the feathers with chordwise and spanwise flaps is shown in Fig.~\ref{fig:res_optcontroller}. The results can be explained in terms of the drag equation commonly seen in literature, $D=\frac{1}{2}\rho v^2C_DS$, where $\rho$ is fluid density, $v$ is velocity, $C_D$ is drag coefficient, and $S$ is area.
The goal is for the feather to produce the greatest asymmetric thrust between its up- and downstrokes. During the upstroke, the feather provides negative thrust, and this is minimized with minimum speed and minimum area from folding. However, since speed aids the folding action of the joints, there is a tradeoff in upstroke speed. For the downstroke, the feather provides positive thrust, which should also be maximized exploiting maximum area and maximum speed. However, depending on the morphology, high speed may result in spanwise bending, which reduces the useful area in producing thrust. 
The hold time after the upstroke helps unfold the feather to its flat, unfolded profile. The hold time after the downstroke helps flatten the slightly over-extended profile of the feather, which requires much less time than the natural unfolding of the feather. 

In Fig.~\ref{fig:res_optcontroller}, two periods of raw thrust data is overlaid with the feather positions for the best and baseline controllers. The prominent characteristics of the thrust profile produced by the best controller labeled in the figure caption. 
In addition to demonstrating the repeatability of the thrust data with the customized experimental setup, we are able to observe varying characteristics in the thrust signals between the best and the baseline controllers. For the feather with chordwise flaps ($WR=7.5$), the optimized controller compared to baseline helps the feather create higher peaks for both negative and positive thrust, which when integrated produce a higher thrust ratio.
This is also true for the feather with spanwise flaps ($LR=2.0$) with the optimized controller producing larger net positive thrust compared to baseline. 
The optimization is useful in finding the best balance between the control parameters that are otherwise difficult to generalize. 


\subsection{Validation of optimization for single feather}

\begin{figure}[tb]
    \centering
    \includegraphics[width=0.9\linewidth]{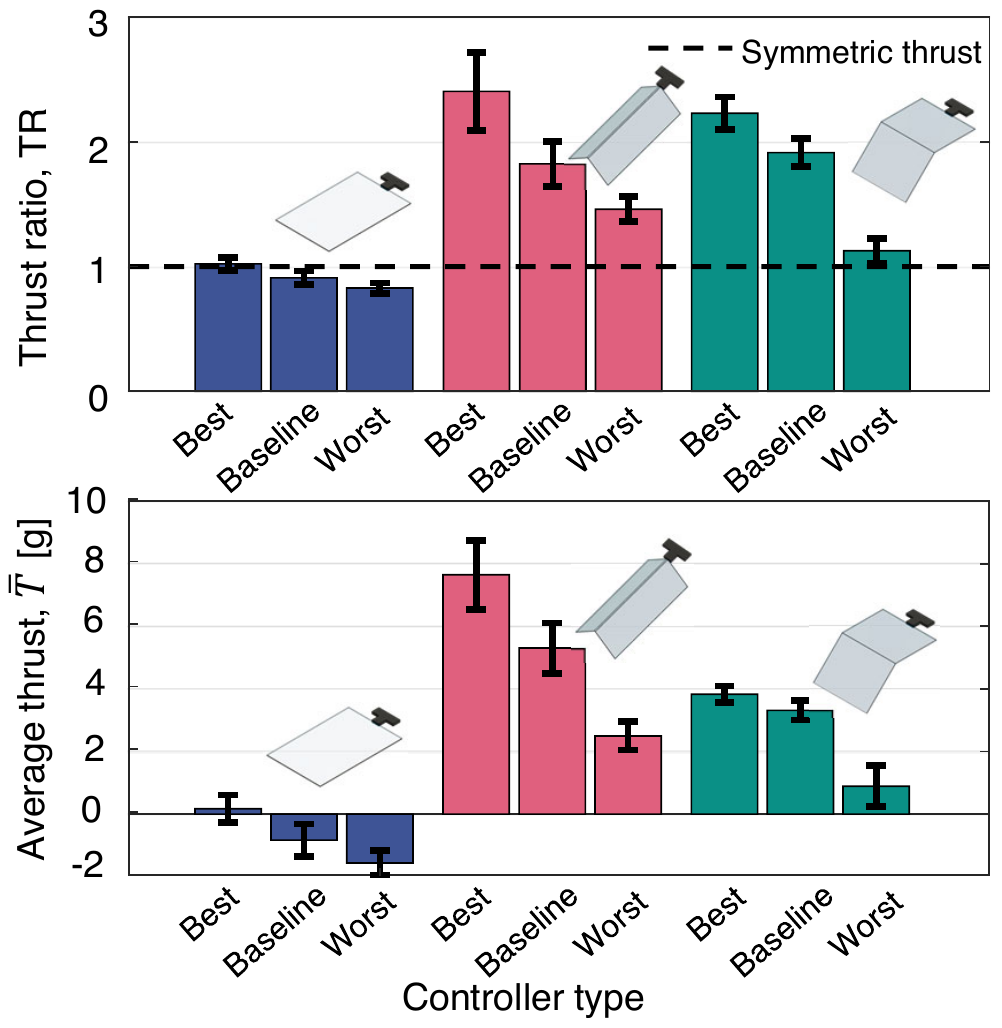}
    \caption{\textbf{Top: }Thrust ratio, $TR$ of all feather designs compared with each other when actuated with the best, baseline (symmetric), and worst controllers. \textbf{Bottom: }Average thrust, $\overline{T}$ of all feather designs actuated with the best, baseline, and worst controllers.}
    \label{fig:res_contvalidation}
\end{figure}

In order to validate the controller optimization for various single feather morphologies, thrust data are collected in the same way as in Sec.~\ref{subsec:asymmetric} (10 sets of data for five periodic cycles per validation experiment) for the best, baseline, and worst controllers. The best and worst controllers are found through automatic experimental optimization with objectives of maximizing and minimizing the thrust ratio, respectively. The baseline controller is identical to that in Sec.~\ref{subsec:asymmetric}. The validation results for a plain feather and two different feathers with flaps ($WR=7.5$ and $LR=2.0$) are shown in Fig.~\ref{fig:res_contvalidation}. 
The plain feather fails to produce any useful asymmetric thrust and even has a tendency to produce more negative thrust ($TR<1$). The results for feathers with asymmetric joints successfully proves the benefits of utilizing passive but variable morphology to produce a net positive thrust. The best controller optimized for maximum thrust ratio provides more than double net positive thrust than negative thrust ($TR>2$) and outperforms the baseline symmetric controller for both feather morphologies with flaps. Furthermore, even when optimizing for minimum thrust ratio, the asymmetric profile is capable of producing more positive thrust ($TR>1$).
As a rough comparison of the absolute thrust production between the feather morphologies, the average thrust calculated as Eq.~\ref{eqn:thrust} is also shown. The average thrust for the feather with spanwise flaps is lower for all controller types, most likely due to the spanwise bending which decreases the total surface area during thrust production.


\subsection{Full robot swimming demonstration}

\begin{figure}[tb]
    \centering
    \includegraphics[width=1\linewidth]{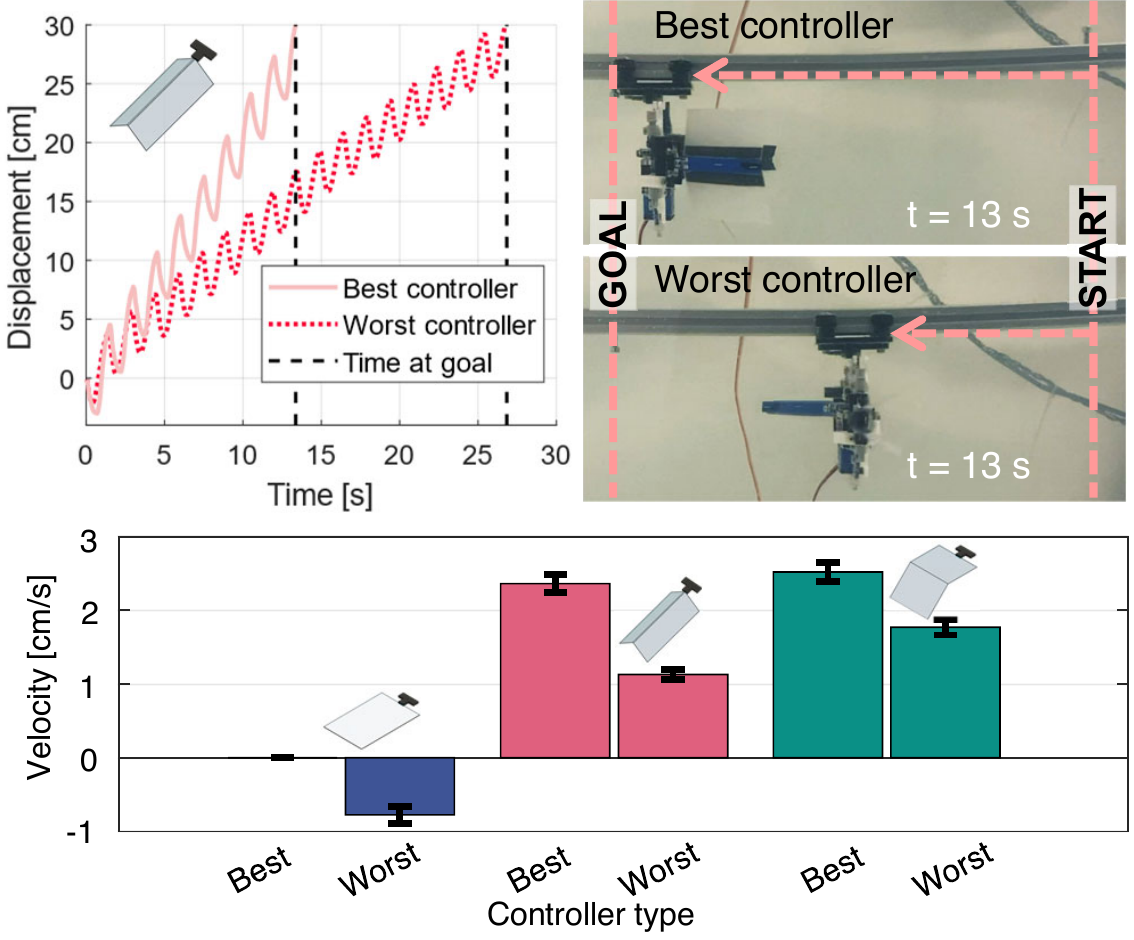}
    \caption{\textbf{Top: } Displacement profile and speed comparison of a full robot fitted with two feathers with chordwise flaps ($WR=7.5$) using the best and worst controllers. \textbf{Bottom: }Velocity comparison of the full robot for all feather designs using the best and worst controllers.}
    \label{fig:res_robot}
    \vspace{-4mm}
\end{figure}

To validate how the thrust profile transfers to a robotic system, swimming experiments are performed with the full robot (Fig.~\ref{fig:fullrobot}) using the best and worst controllers for the plain feather and feathers with chordwise flaps ($WR=7.5$) and with spanwise flaps ($LR=2.0$). Fig.~\ref{fig:res_robot} shows the average velocity results for the different feather morphologies where the robot swam $30\, \mathrm{cm}$ across the aluminum rail across three runs. 
It should be noted that the servo motors used in the single feather experiments and the full robot experiments are different and so there may be variations in the signal. However, in all cases, the robot using the best controller outperforms that with the worst controller.
The full robot demonstration and the displacement profile for the feather with chordwise flaps is also shown (Fig.~\ref{fig:res_robot}) as a clarification for the experiments. The variable morphology in the feather helps the robot achieve a maximum speed of approximately $2.5\, \mathrm{cm/s}$, while the robot with plain flaps fails to move forward.

%% file: 4-discussion.tex
In this work, we explore an intricate relationship between the feather morphology, movement, and the fluid environment where the automated online experimental optimization helps identify a controller to produce the most useful thrust for a given structure. 
The introduction of asymmetry into the design of feathers enables significantly improved directional thrust generation when compared to a feather of the same area.  This benefit requires very little additional `cost' in fabrication or implementation, only exploiting the physicality of the structure.
In the optimization experiments, we demonstrate the sensitivity of these asymmetric flapping structures to different controllers and environments highlighting the need for close co-design of the flapping feather and the controller to enable the flaps to recover.
The thrust profiles generated highlight the complexity of the interactions, and although some hypotheses to why the profile can be presented, further work is required to better understand the physical fluid-structure interactions that lead to these thrust profiles.
Whilst this work motivates the inclusion and design of asymmetric structures in soft swimming robots, further work to incorporate such structures and to develop more robust closed-loop controllers is required to generate free-swimming robotic systems which could leverage this morphological exploitation of fluid-structure interactions.